\documentclass[journal]{IEEEtran}
\usepackage{threeparttable}
\usepackage{graphicx}
\usepackage{amsmath}
\usepackage{amsfonts}
\usepackage{multirow}
\usepackage{multicol}
\usepackage{cite}
\usepackage[numbers]{natbib}
\usepackage{setspace}
\usepackage[ruled,linesnumbered]{algorithm2e}
\ifCLASSINFOpdf

\else

\fi

\begin{document}

\title{BNAS-v2: Memory-efficient and Performance-collapse-prevented Broad Neural Architecture Search}

\author{Zixiang~Ding,~\IEEEmembership{Member,~IEEE},
        Yaran~Chen,~\IEEEmembership{Member,~IEEE},
        Nannan~Li and
        Dongbin~Zhao,~\IEEEmembership{Fellow,~IEEE}% <-this % stops a space
\thanks{Z. Ding, Y. Chen, N. Li and D. Zhao are with the State Key Laboratory of Management and Control for Complex Systems, Institute of Automation, Chinese Academy of Sciences, Beijing 100190, China, and also with the University of Chinese Academy of Sciences, Beijing 100049, China (email : \{dingzixiang2018, chenyaran2013, linannan2017, dongbin.zhao\}@ia.ac.cn).}

%\thanks{This work was supported in part by Youth research fund of the state key laboratory of complex systems management and control No. 20190213 and No. GJHZ1849 International Partnership Program of Chinese Academy of Sciences, and was supported in part by No FA2018111061SOW12 Programe of the Huawei Technologies Co Ltd, and also was supported in part by Noahs Ark Lab, Huawei Technologies.}}

\thanks{This work is supported partly by the National Natural Science Foundation of China (NSFC) under Grants No. 62006226.}
}

%\thanks{Manuscript received April 19, 2019; revised August 26, 2020.}}

%\markboth{IEEE Transactions on Systems, Man, and Cybernetics: Systems, ~2021}%
%{Shell \MakeLowercase{\textit{et al.}}: Bare Demo}

\maketitle


\begin{abstract}

Benefits from Broad Convolutional Neural Network (BCNN), Broad Neural Architecture Search (BNAS) achieves state-of-the-art efficiency in reinforcement learning-based NAS approaches. Particularly, BCNN has two novel characteristics, fast single-step training speed and efficient memory (i.e. larger batch size for architecture search), that all contribute to improve the efficiency of NAS. However, BNAS suffers from the unfair training issue. Consequently, only the former advantage of BCNN plays an important role in BNAS, due to the latter one potentially makes the above issue worse.

In this paper, we propose BNAS-v2 to further improve the efficiency of NAS, embodying both superiorities of BCNN simultaneously. To mitigate the unfair training issue of BNAS, we employ continuous relaxation strategy to make each edge of cell in BCNN relevant to all candidate operations for over-parameterized BCNN construction. Moreover, the continuous relaxation strategy relaxes the choice of a candidate operation as a softmax over all predefined operations. Consequently, BNAS-v2 employs the gradient-based optimization algorithm to simultaneously update every possible path of over-parameterized BCNN, rather than the single sampled one as BNAS. However, continuous relaxation leads to another issue named performance collapse, in which those weight-free operations are prone to be selected by the search strategy. For this consequent issue, two solutions are given: 1) we propose Confident Learning Rate (CLR) that considers the confidence of gradient for architecture weights update, increasing with the training time of over-parameterized BCNN; 2) we introduce the combination of partial channel connections and edge normalization that also can improve the memory efficiency further. Moreover, we denote differentiable BNAS (i.e. BNAS with continuous relaxation) as BNAS-D, BNAS-D with CLR as BNAS-v2-CLR, and partial-connected BNAS-D as BNAS-v2-PC. Experimental results on CIFAR-10 and ImageNet show that 1) BNAS-v2 delivers state-of-the-art search efficiency on both CIFAR-10 (0.05 GPU days that is 4x faster than BNAS) and ImageNet (0.19 GPU days); and 2) the proposed CLR is effective to alleviate the performance collapse issue in both BNAS-D and vanilla differentiable NAS framework.
\end{abstract}

\section{Introduction}

\begin{figure}[!t]
\centering
\includegraphics[width=0.5\textwidth]{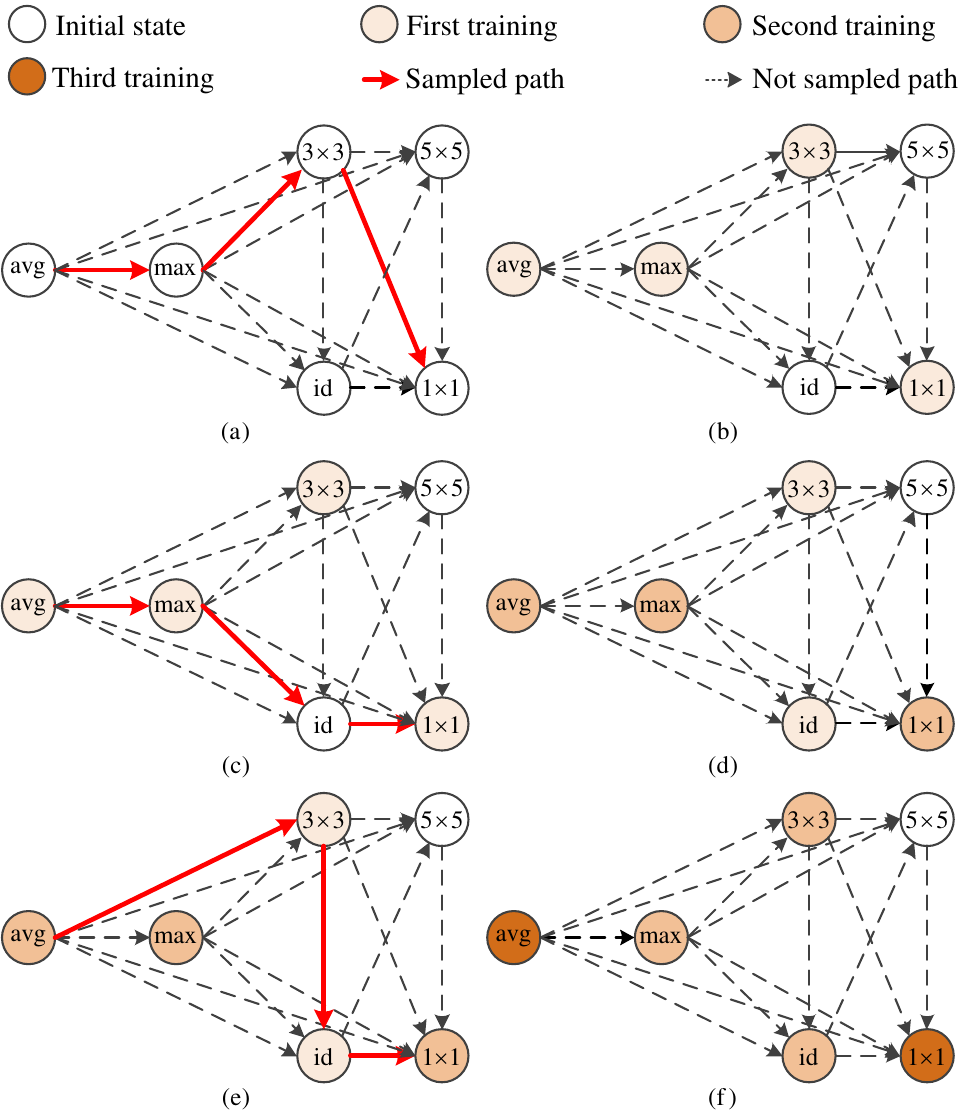}
\caption{Unfair training issue in BNAS. A cell can be treated as a path of directed acyclic graph (DAG). (a) BNAS employs a controller to sample a path (red line) from the DAG with a policy $\pi(\cdot)$; (b) Those operations on the sampled path are trained with a mini-batch training data; (c) The controller samples another path with policy $\pi(\cdot)$; (d) Those operations on the sampled path inherit the weights of previous training step, and are trained with another mini-batch training data; (e) Repeat the above sampling step; (f) Repeat the corresponding training step until an epoch is over. Subsequently, the policy $\pi(\cdot)$ is updated by policy gradient-based RL algorithm that prefers those well trained operations (e.g. $1 \times 1$ convolution in (f)) rather than not well trained ones (e.g. $5 \times 5$ convolution in (f)), due to the former one can obtain higher validation accuracy, i.e. reward. However, if the DAG is trained complete, the convolution with $5 \times 5$ kernel size has larger possibility to obtain higher performance than $1 \times 1$ convolution. Best viewed in color.}
\label{fig::sharing}
\end{figure}

NAS has achieved unprecedented accomplishments in the field of structure design engineering. However, it needs enormous computational requirements, e.g. 22400 GPU days for vanilla NAS \cite{zoph2017neural}. The time consuming issue is mitigated by cell, a micro search space proposed in NASNet \cite{zoph2018learning}. Currently, most of NAS approaches \cite{pham2018efficient,liu2018darts,chen2019progressive,xie2018snas,liang2019darts+,xu2019pc,real2018regularized} are cell-based, where two types of cells are treated as the building blocks of a deep scalable architecture. However, the above deep scalable architecture is time-consuming for both phases of architecture search and evaluation, due to slow single-step training speed and inefficient memory when using multiple cells.

To solve the above issue, \citet{ding2020bnas} proposed BNAS where a broad network paradigm dubbed BCNN was employed as the scalable architecture. Different from previous deep scalable architecture, BCNN could use few cells to deliver competitive performance with broad topology. There were two merits of broad topology compared with the deep one, 1) faster single-step training speed and 2) higher memory efficiency (i.e. architecture search with more training data in a mini-batch), that all conducive to the efficiency improvement of NAS. BNAS delivered state-of-the-art search efficiency of 0.20 GPU days with the combination of Reinforcement Learning (RL) \cite{williams1992simple} and parameters sharing \cite{pham2018efficient}. However, the above optimization strategy of BNAS suffered from the unfair training issue \cite{chu2019fairnas} depicted in Fig. \ref{fig::sharing}. On one hand, the above issue led to the learned architecture with poor performance. On the other hand, the second virtue of BCNN potentially made the above issue worse, due to larger batch size (i.e. higher memory efficiency) would reduce the sample times in a single epoch. As a result, only the virtue of fast single-step training speed of BCNN worked for efficiency improvement of BNAS.

In this paper, we propose memory-efficient and performance-collapse-prevented BNAS named BNAS-v2. Particularly, to mitigate the unfair training issue in BNAS, we employ continuous relaxation strategy to make each edge of cell in BCNN relevant to all candidate operations for over-parameterized BCNN construction. Moreover, the continuous relaxation strategy relaxes the choice of a candidate operation as a softmax over all predefined operations. Consequently, BNAS-v2 employs the gradient-based optimization algorithm to simultaneously update every possible path of over-parameterized BCNN, instead of the single sampled path as BNAS. Furthermore, BNAS-v2 benefits from the memory efficiency of BCNN that contributes to the efficiency improvement and uncertainty reduction for architecture search simultaneously \cite{xu2019pc}. However, performance collapse issue \cite{liang2019darts+} where those weight-free operations are prone to be selected by the search strategy, arises in differentiable BNAS (i.e. BNAS with continuous relaxation) denoted as BNAS-D, resulting in poor performance of the learned architecture. To prevent the consequent issue, we give two solutions for BNAS-D. On one hand, we propose Confident Learning Rate (CLR) that considers the confidence of gradient for architecture weights update, increasing with the training time of over-parameterized BCNN. On the other hand, we introduce the combination of partial channel connections and edge normalization \cite{xu2019pc} that also can further improve the memory efficiency for BNAS-D. Moreover, we denote BNAS-D with CLR as BNAS-v2-CLR, and another partial-connected one as BNAS-v2-PC. Experimental results on CIFAR-10 and ImageNet show that 1) BNAS-v2-CLR and BNAS-v2-PC achieve 2.22x (0.09 GPU days) and 4x (state-of-the-art efficiency of 0.05 GPU days) faster search speed than BNAS on CIFAR-10, respectively; 2) BNAS-v2-PC can directly search on ImageNet with state-of-the-art efficiency of 0.19 GPU days; 3) compared with BNAS, the architectures learned by BNAS-v2 also deliver better performance on CIFAR-10, and competitive performance on ImageNet; and 4) the proposed CLR is effective to prevent the performance collapse issue in not only BNAS-D, but also vanilla differentiable NAS pipeline \cite{liu2018darts}.

The remainder of this paper is organized as follows. We review related work in Section \ref{Related Works}, and introduce the proposed approach in Section \ref{Methodology}. Subsequently, Section \ref{Experiments} shows experiments and corresponding result analysis. At last, we conclude in Section \ref{conclusions}.

\section{Related Work}
\label{Related Works}

%\subsection{Neural Architecture Search}

Hand-crafted neural networks (e.g. ResNet \cite{he2016deep}, GoogleNet \cite{szegedy2015going}) played a predominant role in solving computer vision \cite{zhao2017deep,chen2018multi,wen2020multilabel,wang2020multiscale,chen2020survey,kuang2019feature}, natural language processing  \cite{vaswani2017attention} and other artificial intelligence related tasks \cite{tao2020detection,li2019deep,kamel2019deep,shao2019starcraft} before NAS \cite{zoph2017neural} was proposed. Recent years, NAS achieved unprecedented success in various tasks, e.g. image classification \cite{zoph2018learning,pham2018efficient,liu2018darts,chen2020modulenet,sun2020automatically}, semantic segmentation \cite{liu2019auto}, federated learning \cite{zhu2020multi}.

\begin{figure}[!t]
\centering
\includegraphics[width=0.5\textwidth]{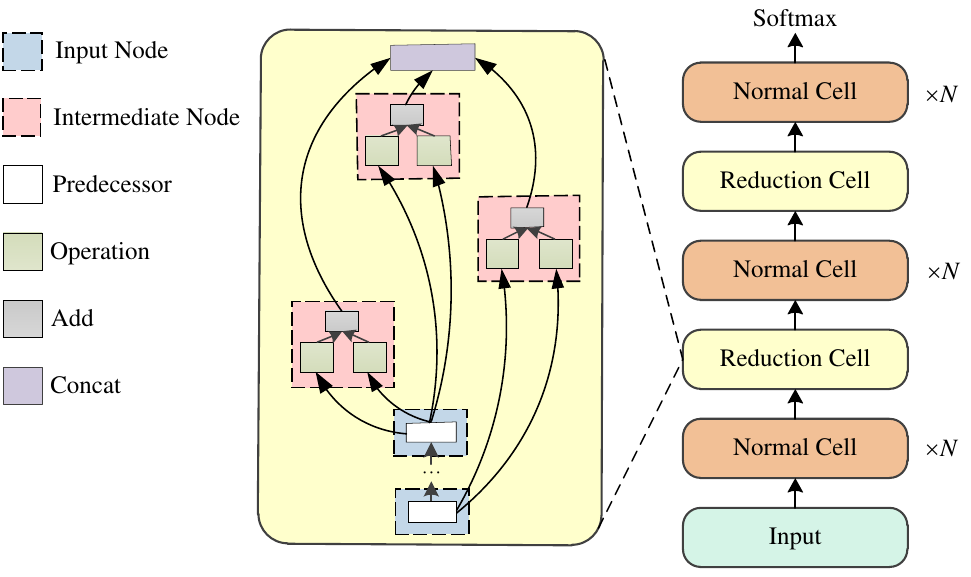}
\caption{Deep scalable architecture. Best viewed in color.}
\label{fig::stack}
\end{figure}

Vanilla NAS \cite{zoph2017neural} greatly suffered from the issue of time consuming. NASNet \cite{zoph2018learning} was proposed to mitigate the above issue, where a micro search space named cell was used to reduce the computational requirements in architecture search phase. Moreover, two types of cells (i.e. normal and reduction cells) were stacked one after another as a deep scalable architecture shown in Fig. \ref{fig::stack}. Subsequently, lots of cell-based NAS approaches were proposed to further improve the efficiency of NAS, e.g. evolutionary algorithm-based LEMONADE \cite{elsken2018efficient}, RL-basd ENAS \cite{pham2018efficient}, gradient-based DARTS \cite{liu2018darts} and a series of variants of DARTS (e.g. SNAS \cite{xie2018snas}, P-DARTS \cite{chen2019progressive}, PC-DARTS \cite{xu2019pc}). DARTS transferred the NAS problem from discrete space to continuous one, and employed gradient-based algorithm to optimize the architecture weights. Furthermore, PC-DARTS adopted the combination of partial channel connections and edge normalization to realize memory-efficient DARTS with novel efficiency improvement.

Inspired by Broad Learning System (BLS) \cite{chen2018universal}, in RL-based BNAS \cite{ding2020bnas}, BCNN employed broad (i.e. shallow) topology to obtain faster single-step training speed and higher memory efficiency than the deep one \cite{zoph2018learning}. Moreover, two variants named BNAS-CCE and BNAS-CCLE with different broad topologies were also proposed for performance promotion. Compared with ENAS \cite{pham2018efficient}, BNAS delivered 2.25x less computation cost of 0.2 GPU day that ranked the best in RL-based NAS pipelines. However, only the advantage of fast single-step training speed contributed to the efficiency improvement of BNAS, due to high memory efficiency (i.e. large batch size) aggravated the unfair training issue of BNAS.

%\subsection{Broad Learning System}
%
%\begin{figure}[!t]
%\centering
%\includegraphics[width=0.45\textwidth]{figure/bls.pdf}
%\caption{Broad learning system.}
%\label{broad learning system}
%\end{figure}
%
%Different from traditional neural networks, the topology of BLS (see Figure \ref{broad learning system}) is broad rather than deep. The combination of topology and optimization algorithm contributed BLS to deliver extreme fast learning speed and competitive performance \cite{chen2017broad}. BLS consisted of two parts: feature nodes and enhancement nodes. On one hand, each group of feature node was fed input data to generate mapped features. On the other hand, enhancement nodes treated the above mapped features as input to provide nonlinear transformation. Finally, all outputs of feature and enhancement nodes were used to compute the output of BLS.
%
%Beyond that, several variants of BLS \cite{chen2018universal} were proposed, e.g. cascade of feature nodes BLS (CFBLS). Some of the above variants were not only broad but also deep, where each group of feature node was connected one after another. Moreover, there were two special variants, 1) CCFBLS: cascade of convolution in feature nodes, and 2) fuzzy BLS: fuzzy model in feature nodes \cite{feng2018fuzzy}. Different from other variants, the components of the above two variants were convolution operation and fuzzy model, instead of neurons.

\begin{table}[!t]
\centering
\begin{threeparttable}[tbq]
\caption{Performance comparison of BNAS and its two variants on CIFAR-10 \cite{ding2020bnas}.}
\label{tab::bnas}
\begin{tabular}{lcccc}
\hline
\multirow{2}{*}{\textbf{Architecture}} & \textbf{Search Cost}$\dag$& \multicolumn{3}{c}{\textbf{Test Error (\%)}$\ddag$}\\
\cline{3-5}
& \textbf{(GPU days)} & \textbf{Small} &  \textbf{Medium} & \textbf{Large}\\
\hline
BNAS      & 0.20 & 3.83 & 3.46 & 2.97 \\
BNAS-CCLE & 0.20 & 3.63 & 3.40 & 2.95 \\
BNAS-CCE  & \textbf{0.19} & \textbf{3.58} & \textbf{3.24} & \textbf{2.88} \\
\hline
\end{tabular}
\footnotesize
\begin{tablenotes}
\item[$\dag$] Searched on a single NVIDIA GTX 1080Ti GPU.
\item[$\ddag$] Each group of cells learned by BNAS or its two variants is stacked to small (about 0.5 millions), medium (about 1.1 millions) and large-size (about 4 millions) models.
\end{tablenotes}
\end{threeparttable}
\end{table}

\begin{figure}[!t]
\centering
\includegraphics[width=0.47\textwidth]{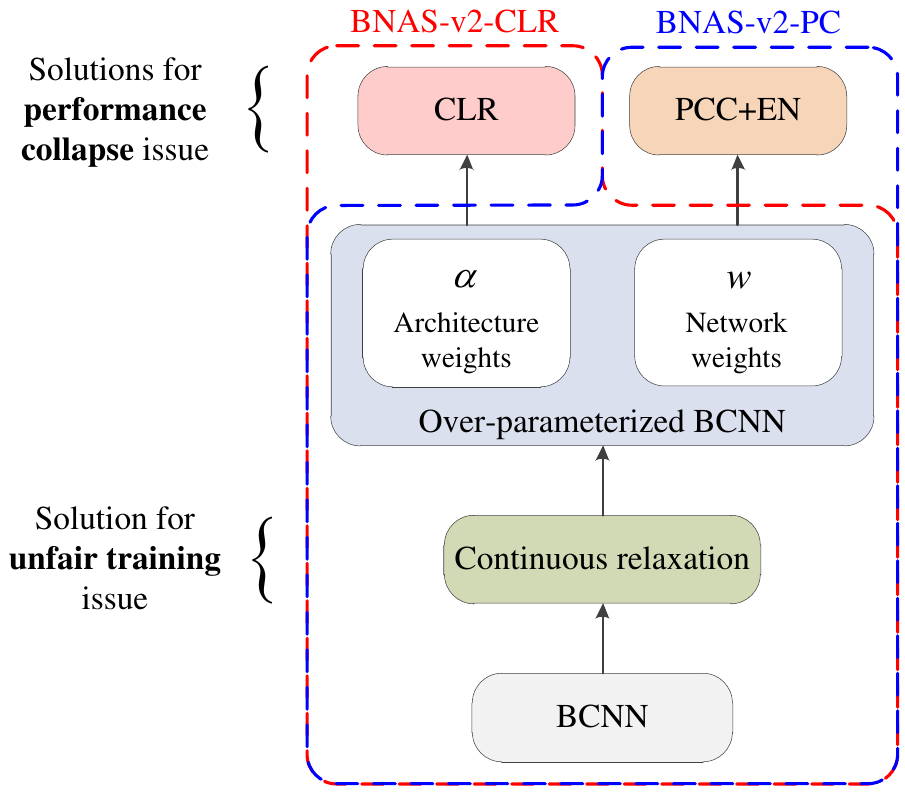}
\caption{Overview of BNAS-v2. Continuous relaxation is employed to mitigate the aforementioned unfair training issue, so as to BNAS-2 can take full advantages of BCNN, i.e. fast single-step training speed and efficient memory. Consequently, performance collapse issue arises. To prevent the consequent issue, we propose CLR and introduce the combination of partial channel connections (PCC) and edge normalization (EN). Moreover, the later one is able to improve the memory efficiency of BNAS-v2 further so that BNAS-v2-PC can directly discover architecture on ImageNet. Best viewed in color.}
\label{fig::overview}
\end{figure}

\section{Memory-efficient and Performance-collapse-prevented BNAS}
\label{Methodology}
As aforementioned, there are three types of BCNNs in BNAS with different performances shown in TABLE \ref{tab::bnas} \cite{ding2020bnas}. Obviously, BNAS-CCE delivers the best performance with respect to both efficiency and accuracy. As a result, we treat the BCNN employed in BNAS-CCE as the broad scalable architecture for BNAS-v2.

The overview of BNAS-v2 is shown in Fig. \ref{fig::overview}. To mitigate the unfair training issue, we employ continuous relaxation for over-parameterized BCNN construction, so that gradient-based optimization algorithm can be employed to simultaneously update every possible path of over-parameterized BCNN, rather than the single sampled one as BNAS. Furthermore, two solutions are given to prevent the consequent issue of continuous relaxation, i.e. performance collapse. On one hand, we propose Confident Learning Rate (CLR) that considers the confidence of gradient for architecture weights update increasing with the training time of over-parameterized BCNN. On the other hand, we introduce the combination of partial channel connections and edge normalization \cite{xu2019pc} that also can improve the memory efficiency further, so that BNAS-v2-PC can directly discover architecture on ImageNet.

\subsection{Preliminaries: Broad Convolutional Neural Network}

\begin{figure}[!t]
\centering
\includegraphics[width=0.45\textwidth]{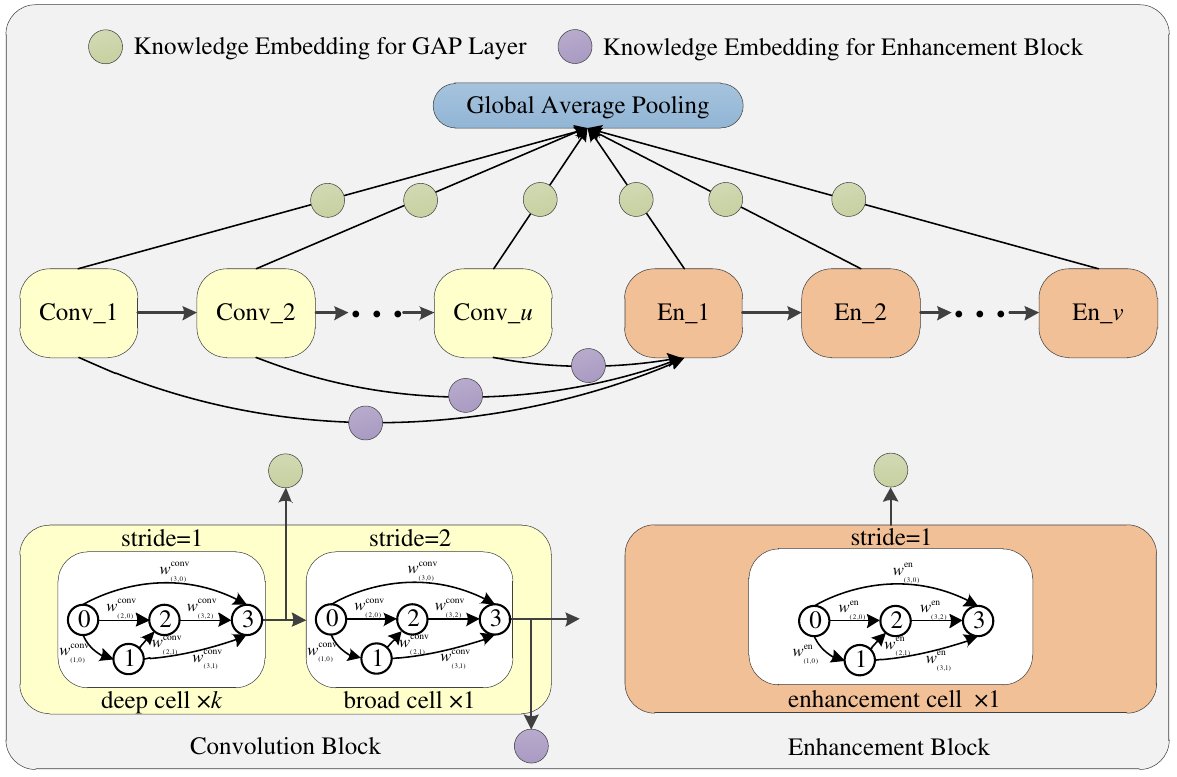}
\caption{Broad Convolutional Neural Network. There are four components: \emph{convolution block} for feature extraction, \emph{enhancement block} for feature enhancement, \emph{knowledge embedding} for feature significance control and \emph{multi-scale feature fusion} for the guarantee of high-performance BCNN. Best viewed in color.}
\label{fig::bcnn}
\end{figure}

We show the structure of cell-based BCNN in Fig. \ref{fig::bcnn}. There are four components: \emph{convolution block}, \emph{enhancement block}, \emph{knowledge embedding} and \emph{multi-scale feature fusion}. \emph{Convolution block} employs $k$ deep cells (stride=1) and a single broad cell (stride=2) for deep and broad feature extraction, respectively. The first \emph{enhancement block} treats the outputs of every broad cell as input for representation enhancement. All \emph{enhancement block}s (stride=1) are stacked one after another. \emph{Knowledge embedding} is inserted into BCNN to control the significance of the output of each \emph{convolution block} for the Global Average Pooling (GAP) layer and the first \emph{enhancement block}. \emph{Multi-scale feature fusion} realizes that GAP can fuse multi-scale features to more comprehensive representation for high classification accuracy with broad (i.e. shallow) topology.

%Due to the novel design of BCNN, BNAS or BNAS-v2 is able to employ few cells for architecture search with high efficiency and keep satisfactory performance in the architecture estimation phase.

\subsection{Continuous Relaxation of BCNN for Unfair Training}
\label{pipeline}

Each cell in BCNN (i.e. deep, broad or enhancement cell) is a directed acyclic graph containing $N$ nodes: 2 input nodes $\{x_{(0)},x_{(1)}\}$, $N-3$ intermediate nodes $\{x_{(2)},\dots,x_{(N-2)}\}$, and a single output node $x_{(N-1)}$. Each intermediate node $x_{(i)}$ is a set of feature maps obtained by operations $o_{(i,j)}(\cdot)$, which are chosen from a predefined search space $\mathcal{O}$ consisting of multiple candidate operations (e.g. convolution, pooling) and used to transform $x_{(j)}$. Hence, each intermediate node can be represented by
\begin{equation}
x_{(i)} = \sum_{j<i}o_{(i,j)}(x_{(j)}).
\label{intermediate output}
\end{equation}
The output of cell is obtained by concatenating the outputs of all intermediate nodes.

In order to alleviate the unfair training issue, we make each intermediate node relevant to all candidate operations that treat the output of all predecessor nodes as input for over-parameterized BCNN construction, so that the optimization algorithm of BNAS-v2 can simultaneously update every possible path (i.e. cell) rather than the single sampled one as BNAS. Inspired by the strategy of continuous relaxation \cite{liu2018darts}, we relax edge $(i,j)$ of each cell for BCNN by
\begin{equation}
f_{(i,j)}(x_{(j)}) = \sum_{o \in \mathcal{O}}\frac{{\rm exp}(\alpha_{(i,j)}^o)}{\sum_{o' \in \mathcal{O}}{\rm exp}(\alpha_{(i,j)}^{o'})}o(x_{(j)}),
\label{relaxation}
\end{equation}
where, operation $o(x_{(j)})$ is weighted by a hyper-parameter $\alpha_{(i,j)}^o$ of dimension $|\mathcal{O}|$. Subsequently, BNAS is developed to differentiable BNAS denoted as BNAS-D so that gradient-based algorithm can be employed for parameter optimization. In this paper, we denote $\alpha$ as architecture weights, and the parameters of operations $w$ as network weights.

\subsection{Two Solutions to Prevent Performance Collapse}

After continuous relaxation, BNAS-D is able to take two advantages of BCNN, so that it is memory-efficient. However, the performance of BCNN learned by BNAS-D is poor due to a consequent issue of continuous relaxation. The above issue is denoted as performance collapse \cite{liang2019darts+} that make optimization algorithm prefer to choose those weight-free operations, e.g. skip connection, pooling. To mitigate this consequent issue, two solutions are provided and shown as below.

\subsubsection{Confident Learning Rate}

\begin{figure}[t]
\centering
\includegraphics[width=0.42\textwidth]{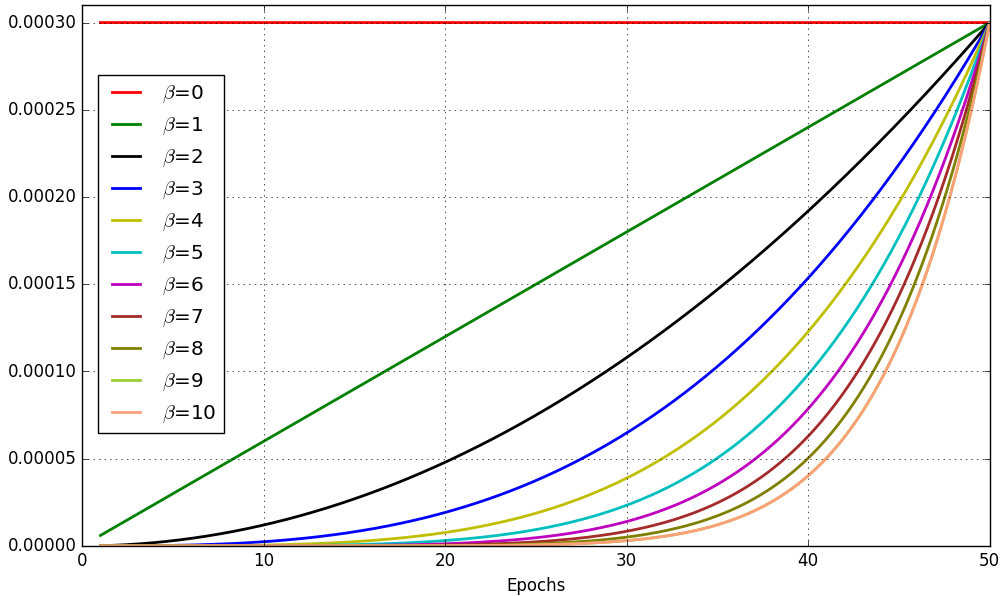}
\caption{The curves of CLR under various confidence factors with default architecture learning rate 0.0003 for differentiable NAS approaches. Best viewed in color.}
\label{fig::clr}
\end{figure}

Compared with those weight-equipped operations (e.g. convolution), weight-free operations tend to obtain larger weights before the weights of over-parameterized BCNN are well optimized \cite{xu2019pc}. In other words, the confidence of gradient obtained from over-parameterized BCNN should increase with the training time for architecture weights update, so that CLR with respect to the number of current epoch is proposed as
\begin{equation}
lr_{conf}(t) = (\frac{t}{T})^{\beta} \times lr_{\alpha},
\label{clr}
\end{equation}
where, $t$ denotes the current epoch from 1 to the maximum $T$, $\beta$ represents the confidence factor whose value is directly proportional to the confidence of early over-parameterized BCNN, and $lr_{\alpha}$ is the initial learning rate for architecture weights. For intuitive comprehension, we plot the curves of CLR under different confidence factors in Fig. \ref{fig::clr} where $lr_{\alpha}=0.0003$ is the default setting for differentiable NAS framework \cite{liu2018darts}.

How to determine the value of confidence factor $\beta$ is an intractable problem. With $\beta$ increasing, more epochs of early training process are involved to freeze the architecture weights, similar to the strategy of warmup used in PC-DARTS (i.e. training architecture weights after 15 epochs). We make a criterion for $\beta$ determination as follows.

\begin{figure}[t]
\centering
\includegraphics[width=0.345\textwidth]{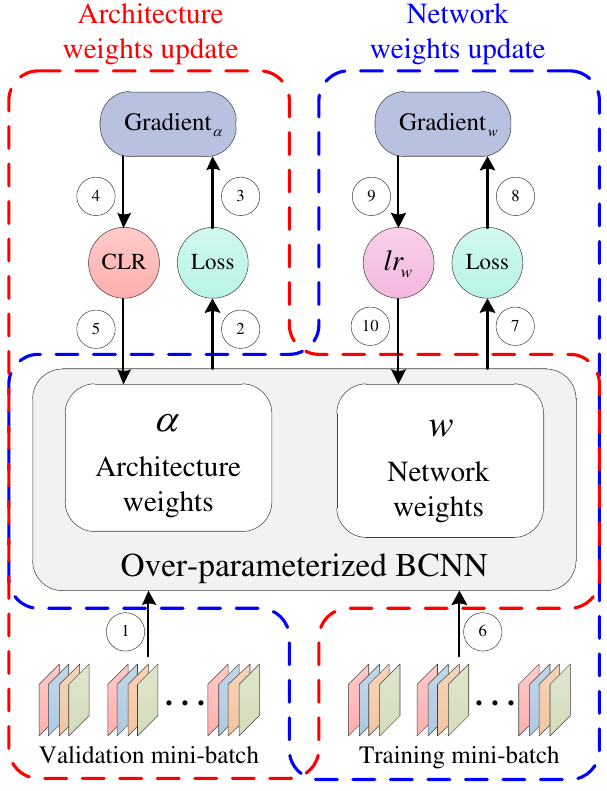}
\caption{Overview for BNAS-v2-CLR optimization. For architecture search, there is a loop as below. 1) \emph{Architecture weights update}: \textcircled{\scriptsize{1}} An validation mini-batch is fed into the broad scalable architecture, to \textcircled{\scriptsize{2}} obtain the loss for \textcircled{\scriptsize{3}} computing the gradient with respect to architecture weights. The proposed CLR is applied for \textcircled{\scriptsize{4}} building confident gradient, i.e. $grad_{conf}=lr_{conf} * \nabla_{\alpha}\mathcal{L}_{val}(w-\xi\nabla_{w}\mathcal{L}_{train}(w,\alpha),\alpha)$, for \textcircled{\scriptsize{5}} architecture weights update. 2) \emph{Network weights update}: \textcircled{\scriptsize{6}} A training mini-batch is treated as the input of broad scalable architecture. Subsequently, \textcircled{\scriptsize{7}} the loss is obtained and \textcircled{\scriptsize{8}} the gradient with regard to network weights is computed. Finally, \textcircled{\scriptsize{9}} the product of learning rate $lr_{w}$ and the gradient with respect to $w$, i.e. $lr_{w} * \nabla_{w}\mathcal{L}_{train}(w,\alpha)$ are employed for \textcircled{\scriptsize{10}} network weights update. Best viewed in color.}
\label{fig::trainingclr}
\end{figure}

\textbf{Criterion 1}: \emph{The optimal confidence factor should make the architecture weights start to be updated at about 15-th epoch for BNAS-D.}

\begin{figure*}[t]
\centering
\includegraphics[width=0.96\textwidth]{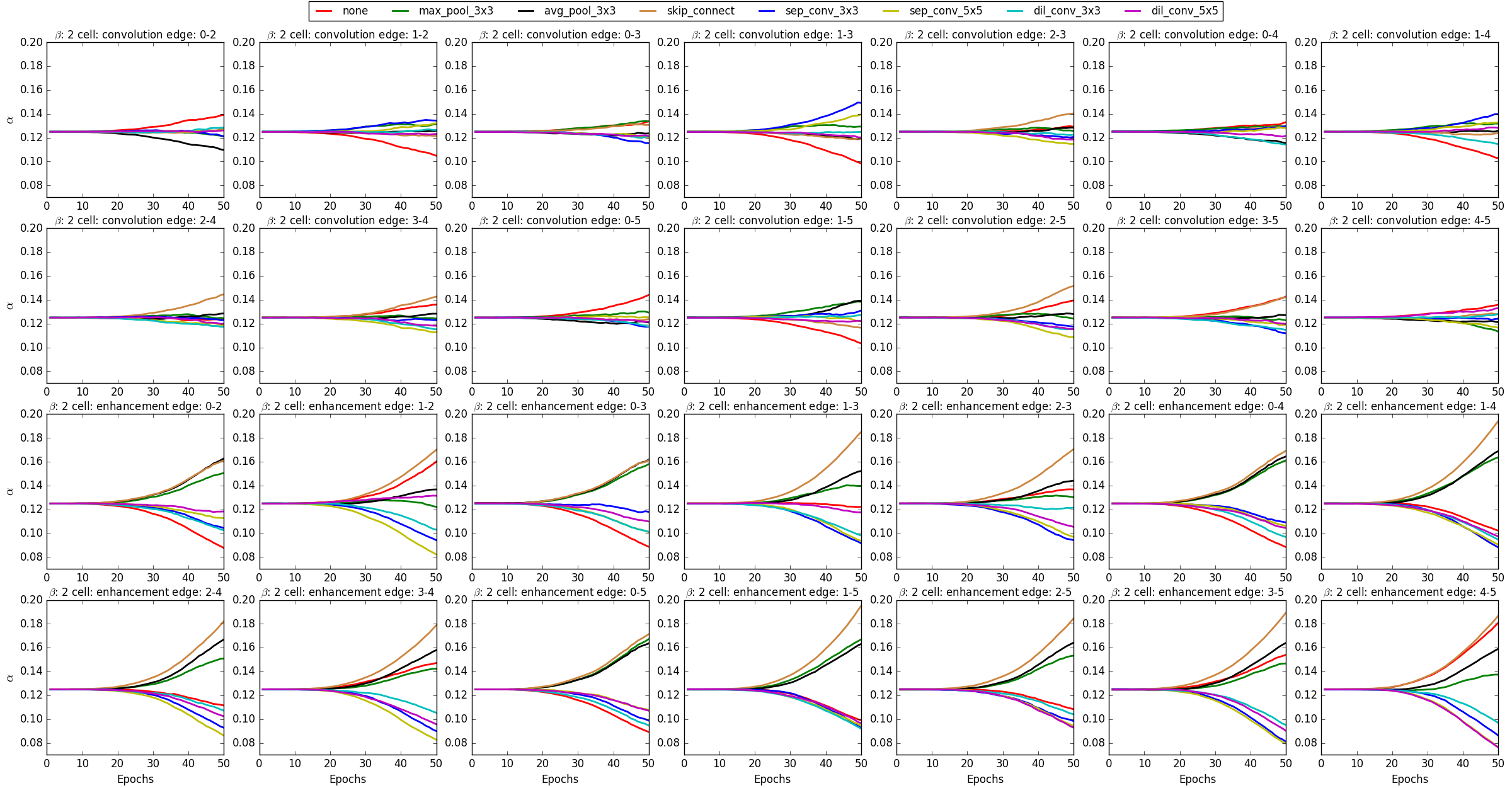}
\caption{Performance collapse issue of convolution (subgraph 1 to 14) and enhancement (subgraph 15 to 28) cells in BNAS-v2-CLR ($\beta=2$) with \emph{skip connection}. Almost all edges choose \emph{skip connection} as operations in enhancement cell that leads to poor-performance BCNN. Best viewed in color.}
\label{collapse}
\end{figure*}

\subsubsection{Partial Channel Connections and Edge Normalization}

For the connection from $x_{(j)}$ to $x_{(i)}$, partial channel connections (PCC) \cite{xu2019pc} feeds partial channels into $|\mathcal{O}|$ operations, and copies others to the output directly. Consequently, the continuous relaxation of BNAS-v2-PC can be computed by
\begin{equation}
\begin{aligned}
f^{PC}_{(i,j)}(x_{(j)}; M_{(i,j)}) &= \sum_{o \in \mathcal{O}}\frac{{\rm exp}(\alpha_{(i,j)}^o)}{\sum_{o' \in \mathcal{O}}{\rm exp}(\alpha_{(i,j)}^{o'})}o(M_{(i,j)} * x_{(j)})\\
&+(1-M_{(i,j)})*x_{(j)},\\
\end{aligned}
\label{eq::pc}
\end{equation}
where, $M_{(i,j)}$ denotes channels sampling mask whose values are chosen from 0 (masked channels) and 1 (selected channels), $M_{(i,j)} * x_{(j)}$ represents the chosen channels and $(1-M_{(i,j)})*x_{(j)}$ computes the masked one. Benefits from PCC, the memory usage of BNAS-v2-PC is reduced greatly, so that larger batch size can be set and higher efficiency can be achieved compared with BNAS-v2-CLR.

After the strategy of PCC, only small part of input channels are fed into the operation mixture that alleviates performance collapse issue \cite{xu2019pc}. However, PCC also causes undesired fluctuation in the architecture learning process. To mitigate the above problem, we employ edge normalization where edge $(i,j)$ is weighted by another hyper-parameter denoted as $\gamma_{(i,j)}$, so that $x_{(i)}$ can be computed as
\begin{equation}
x_{(i)}^{PC}=\sum_{j<i}\frac{{\rm exp}(\gamma_{(i,j)})}{\sum_{j'<i}{\rm exp}(\gamma_{(i,j')})}\cdot f_{(i,j)}(x_{(j)}).
\label{xpc}
\end{equation}
The operation $o_{(i,j)}$ of final architecture is determined by

\begin{equation}
o_{(i,j)}=\mathop{{\rm argmax}}\limits_{o \in \mathcal{O}}\frac{{\rm exp}(\alpha_{(i,j)}^o)}{\sum_{o' \in \mathcal{O}}{\rm exp}(\alpha_{(i,j)}^{o'})} \cdot \frac{{\rm exp}(\gamma_{(i,j)})}{\sum_{j'<i}{\rm exp}(\gamma_{(i,j')})}.
\label{product}
\end{equation}

\subsection{Optimization Strategies for BNAS-v2}

Benefits from continuous relaxation, gradient-based algorithm can be employed for architecture optimization. We propose different strategies, BNAS-v2-CLR and BNAS-v2-PC, as solutions to prevent performance collapse.

\subsubsection{Architecture Search for BNAS-v2-CLR}

We show the training process in Fig. \ref{fig::trainingclr}. Similar to vanilla differentiable NAS pipeline \cite{liu2018darts}, BNAS-v2-CLR updates $\alpha$ and $w$ by descending $\nabla_{w}\mathcal{L}_{train}(w,\alpha)$ and $lr_{conf}*\nabla_{\alpha}\mathcal{L}_{val}(w-\xi\nabla_{w}\mathcal{L}_{train}(w,\alpha),\alpha)$, respectively. Moreover, $\mathcal{L}$ represents loss function, and $\xi$ is used to control the approximation order by
\begin{equation}
\left\{
\begin{aligned}
&\xi = 0, \quad 1st \; order\\
&\xi > 0, \quad 2nd \; order\\
\end{aligned}
\right..
\end{equation}

\subsubsection{Architecture Search for BNAS-v2-PC}

\begin{algorithm}[!t]
\caption{Architecture Search for BNAS-v2-PC}
\label{algorithm}
For each edge $(i,j)$, use \eqref{eq::pc} and \eqref{xpc} for continuous rela- xation as $f^{PC}_{(i,j)}(x_{(j)}^{PC}; M_{(i,j)})$ parameterized by $\alpha_{(i,j)}$ and $\gamma_{(i,j)}$\;
\While{not converged}
{Optimize $w$ by descending $\nabla_{w}\mathcal{L}_{train}(w,\alpha,\gamma)$\;
Optimize $\alpha$ and $\gamma$ by descending $\nabla_{\alpha}\mathcal{L}_{val}(w-\xi\nabla_{w}\mathcal{L}_{train}(w,\alpha,\gamma),\alpha,\gamma)$\;}
Determine $o_{(i,j)}$ by \eqref{product}.
\end{algorithm}

The proposed architecture optimization algorithm is shown in \textbf{Algorithm \ref{algorithm}}. There are several differences between the optimization strategies of BNAS-v2-CLR and BNAS-v2-PC as follows.
\begin{itemize}
  \item BNAS-v2-PC employs the strategy of PCC and EN to construct partial-connected over-parameterized BCNN;
  \item The architecture learned by BNAS-v2-PC is determined by $\alpha$ and $\gamma$.
\end{itemize}

\section{Experiments}
\label{Experiments}

\subsection{Datasets and Implementation Details}

\subsubsection{Datasets}
Similar to previous works \cite{zoph2018learning,xie2018snas,chen2019progressive}, CIFAR-10 \cite{krizhevsky2009learning} and ImageNet \cite{russakovsky2015imagenet} are also selected for BNAS-v2. There are 60K images with spatial resolution of $32\times32$ in CIFAR-10 \cite{krizhevsky2009learning}, where 50K for training and 10K for test. A list of standard methods are applied for preprocessing of CIFAR-10, e.g. randomly flipping and cropping. ImageNet \cite{russakovsky2015imagenet} is a popular dataset for large scale image classification task. There are about 1.3M images with various spatial resolutions in ImageNet, that are near equally distributed over 1000 object categories. Similarly, a series of data preprocessing techniques are applied to ImageNet. Following previous works \cite{zoph2018learning,liu2018darts,chen2019progressive,xie2018snas,liang2019darts+,xu2019pc}, we reshape the size of original images of ImageNet to $224\times224$.

\subsubsection{Implementation Details}

BNAS-v2-CLR uses CIFAR-10 for architecture search and ImageNet for transferability verification. Differently, we employ BNAS-v2-PC to directly search on not only CIFAR-10 but also ImageNet. In previous works \cite{liu2018darts,xu2019pc}, search space $\mathcal{O}$ consists of 8 operations, i.e. \emph{separable convolution} with $3\times3$ and $5\times5$ kernels, \emph{dilated separable convolution} with $3\times3$ and $5\times5$ kernels, \emph{max pooling} and \emph{average pooling} with $3\times3$ kernel, \emph{skip connection}, and \emph{zero} (\emph{none}). We still utilize the above candidate operations as search space for BNAS-v2-PC. Differently, \emph{skip connection} is removed from the search space of BNAS-v2-CLR, due to CLR can alleviate the collapse issue in convolution cell rather than enhancement cell. We visualize the architecture weights with respect to each edge of convolution and enhancement cells in Fig. \ref{collapse}. Obviously, the proposed CLR can not make the \emph{skip connection} (i.e. the blue line) out of predominance, especially for the enhancement cell. Moreover, full \emph{skip connection}-consisted enhancement cell dose not work for broad scalable architecture. As described in DARTS+ \cite{liang2019darts+}, the first cell employs fresh images as input and the input of last one is mixed with lots of noises, so that the enhancement cell suffers from the collapse issue worse than the convolutional one.

\begin{figure}[t]
\centering
\includegraphics[width=0.48\textwidth]{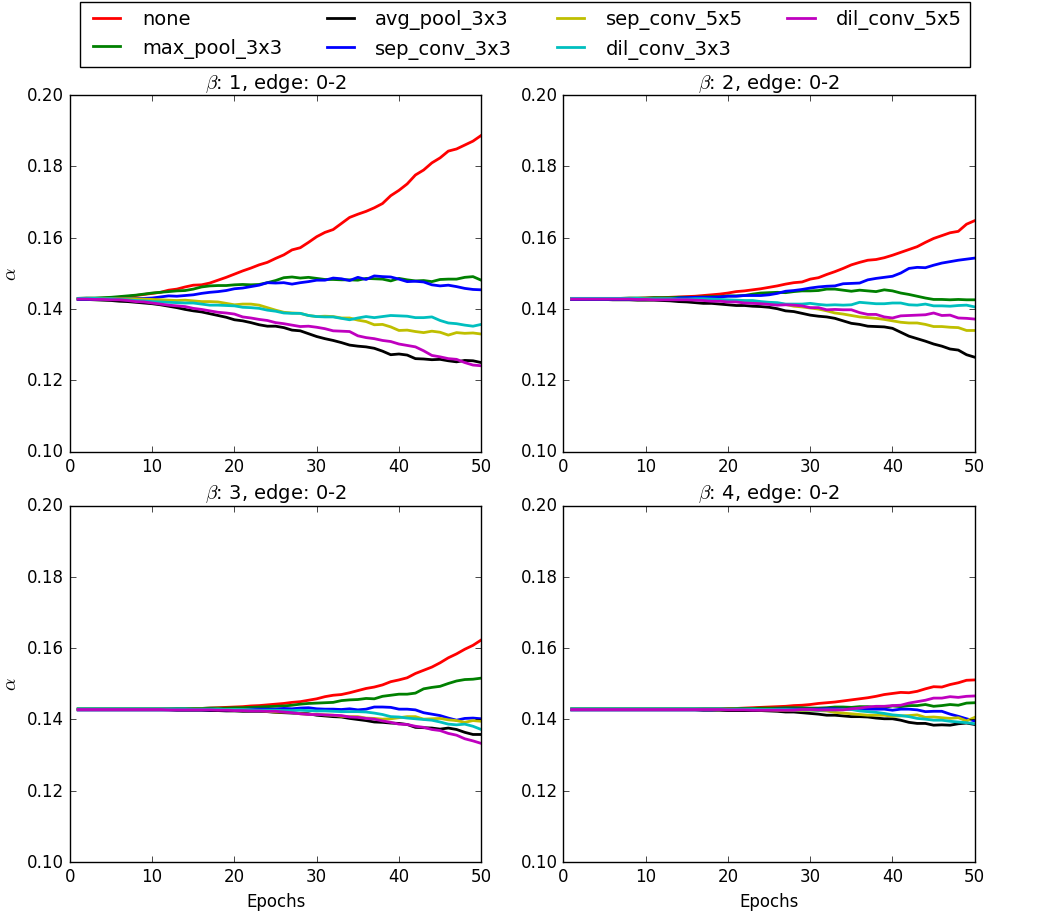}
\caption{The architecture weights of each operation on the shallowest edge for BNAS-v2-CLR under various confidence factors. Best viewed in color.}
\label{power_determination}
\end{figure}

\begin{figure}[t]
\centering
\includegraphics[width=0.43\textwidth]{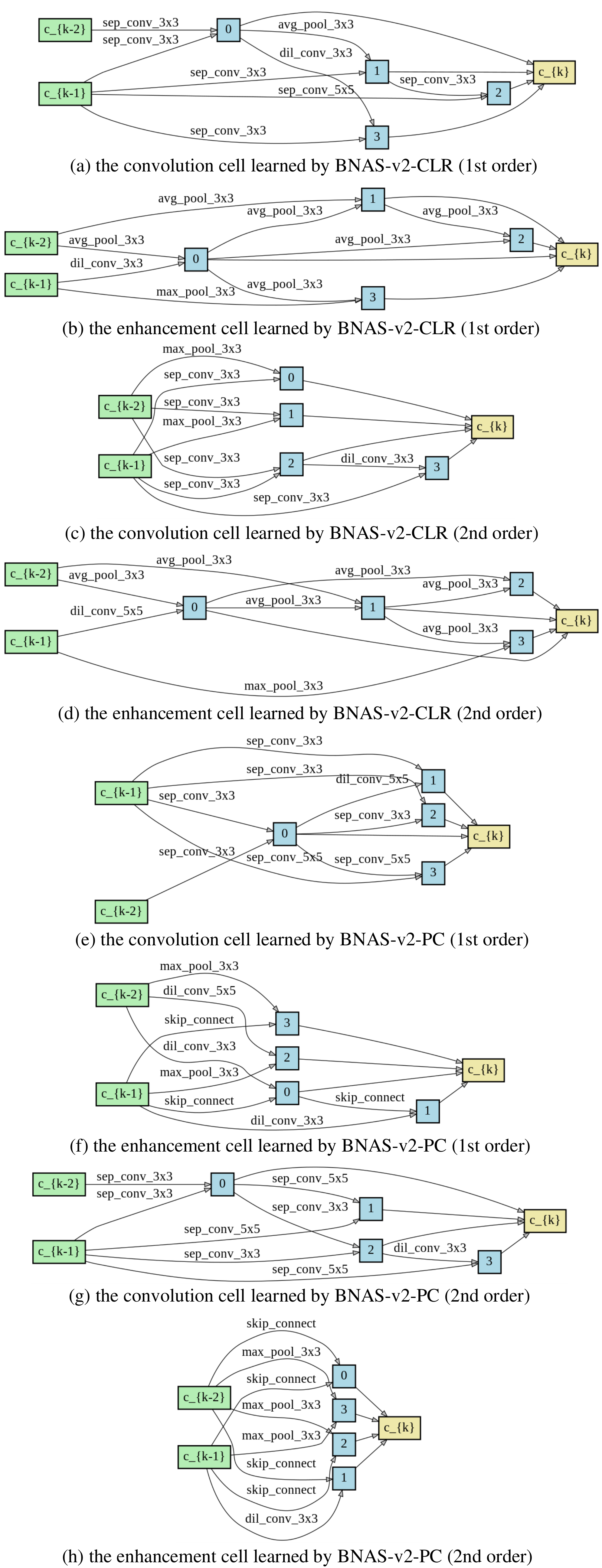}
\caption{The architectures learned by BNAS-v2 on CIFAR-10.}
\label{cells}
\end{figure}

\subsection{Confidence Factor Determination}
\label{Confidence Factor Determination}

In order to determine optimal $\beta$ under \textbf{Criterion 1}, we set it from 1 to 4 for architecture search. The architecture weights $\alpha$ with respect to the shallowest edge (connecting the first input node $x_{(0)}$ and the first intermediate node $x_{(2)}$) is chosen as the index for confidence factor determination. We show the experimental results for BNAS-v2-CLR in Fig. \ref{power_determination}.

\begin{table*}[!t]
\centering
\begin{threeparttable}[tbq]
\caption{Comparison of the proposed BNAS-v2 and other state-of-the-art NAS approaches on CIFAR-10.}
\label{tab::cifar10}
\begin{tabular}{lcccccc}
\hline
\multirow{2}{*}{\textbf{Architecture}} & \textbf{Error}& \textbf{Params}& \textbf{Search Cost} & \multirow{2}{*}{\textbf{Search Method}} &\textbf{Number}&\multirow{2}{*}{\textbf{Topology}}\\
& \textbf{(\%)} & \textbf{(M)}   & \textbf{(GPU days)} & & \textbf{of Cells} \\
\hline
LEMONADE \cite{elsken2018efficient}    & 3.05 & 4.7  & 80  & evolution & - & deep\\
DARTS (1st order) \cite{liu2018darts} & 3.00 $\pm$ 0.14 & 3.3  & 0.45 $\dag$ & gradient-based & 20 & deep \\
DARTS (2nd order) \cite{liu2018darts}& 2.76 $\pm$ 0.09 & 3.3  & 1.50 $\dag$ & gradient-based & 20 & deep  \\
SNAS + mild constraint \cite{xie2018snas}  & 2.98 & 2.9  & 1.5 & gradient-based & 20 & deep \\
SNAS + moderate constraint \cite{xie2018snas}  & 2.85 $\pm$ 0.02 & 2.8  & 1.5 & gradient-based & 20 & deep \\
SNAS + aggressive constraint \cite{xie2018snas}  & 3.10 $\pm$ 0.04 & \textbf{2.3}  & 1.5 & gradient-based & 20 & deep \\
P-DARTS \cite{chen2019progressive} & \textbf{2.50} & 3.4  & 0.30 & gradient-based & 20 & deep \\
PC-DARTS (1st order) \cite{xu2019pc} & 2.57 $\pm$ 0.07 & 3.6  & 0.10 & gradient-based & 20 & deep \\
PC-DARTS (2nd order) \cite{xu2019pc} & - & - & OOM $\ddag$ & gradient-based & - & - \\
ENAS \cite{pham2018efficient} & 2.89 & 4.6  & 0.45 & RL & 17 & deep \\
BNAS \cite{ding2020bnas}      & 2.97 & 4.7 & 0.20 & RL & \textbf{5} & broad \\
BNAS-CCLE \cite{ding2020bnas} & 2.95 & 4.1 & 0.20 & RL & \textbf{5} & broad \\
BNAS-CCE \cite{ding2020bnas}  & 2.88 & 4.8 & 0.19 & RL & 8 & broad \\
\hline
BNAS-v2-CLR (1st order) (ours) & 2.67 $\pm$ 0.12 & 3.3 & 0.09 & gradient-based & 8 & broad\\
BNAS-v2-CLR (2nd order) (ours) & 2.80 $\pm$ 0.09 & 3.2 & 0.19 & gradient-based & 8 & broad \\
BNAS-v2-PC (1st order) (ours)  & 2.79 $\pm$ 0.19 & 3.7 & \textbf{0.05} & gradient-based & 8 & broad\\
BNAS-v2-PC (2nd order) (ours)  & 2.77 $\pm$ 0.09 & 3.5 & 0.09 & gradient-based & 8  & broad\\
\hline
\end{tabular}
\footnotesize
\begin{tablenotes}
\item[$\dag$] Obtained by DARTS using the code publicly released by the authors at https://github.com/quark0/darts on a single NVIDIA GTX 1080Ti GPU.
\item[$\ddag$] Obtained by PC-DARTS using the code publicly released by the authors at https://github.com/yuhuixu1993/PC-DARTS with default setting for 1st order approximation of batch size 256 on a single NVIDIA GTX 1080Ti GPU.
\end{tablenotes}
\end{threeparttable}
\end{table*}

For the first case of BNAS-v2-CLR, $\alpha$ starts to be updated before 10-th epoch. For $\beta=3$ and $\beta=4$, the starting epochs of weights update are both larger than 20. The case of $\beta=2$ satisfies \textbf{Criterion 1}, i.e. starting to train architecture weights from about 15-th epoch. Consequently, $\beta$ is set to 2 for the following experiments with regard to BNAS-v2-CLR. Furthermore, we find an interesting phenomenon from Fig. \ref{power_determination}, where the \emph{none} operation (i.e. red line) always achieves the largest weight when using various confidence factors. The above phenomenon indicates that the proposed CLR modifies the tendency of weight-equipped operations in later training epochs, rather than imposing small architecture weight on weight-free operations in early training phase.

%Different from B-DARTS, we set the confidence factor $\beta$ to 4 instead of 2 for DARTS. In the search phase of DARTS, there are more cells in the over-parameterized model than B-DARTS. Hence, DARTS needs more training data to optimize network parameters well.

\subsection{Experiments on CIFAR-10}

%\begin{figure}[!t]
%\centering
%\includegraphics[width=0.502\textwidth]{./figure/cells_pc.pdf}
%\caption{The architectures learned by BNAS-v2-PC on CIFAR-10 and ImageNet.}
%\label{cells_pc}
%\end{figure}

\subsubsection{Experimental Settings}
In the architecture search phases of BNAS-v2-CLR and BNAS-v2-PC, there are many identical experimental details as below. The over-parameterized BCNN consists of 3 cells (2 broad cells and 1 enhancement cell), where each one contains 2 input nodes, 4 intermediate nodes and 1 output node. We set the number of initial input channels to 16 and train the over-parameterized BCNN for 50 epochs. The split portion of proxy dataset is set to 0.5, i.e. two subsets, each one with 25K training data of CIFAR-10, that are used for training the over-parameterized BCNN and architecture weights, respectively. The SGD optimizer \cite{loshchilov2016sgdr} is employed to optimize the network weights $w$, with dynamic learning rate (annealed down to zero following a cosine schedule without restart), momentum 0.9, weight decay $3\times10^{-4}$. For architecture weights, we use zero initialization to generate $\alpha$ for both convolution and enhancement cells. Moreover, we utilize Adam \cite{kingma2015adam} with momentum (0.5, 0.999) and weight decay $10^{-3}$, as the optimizer to update $\alpha$. We run four repeated experiments of architecture search for BNAS-v2-CLR and BNAS-v2-PC, and choose the best performing architecture as the optimal one that will run three times in architecture evaluation phase. There are also some differences between the experimental settings of BNAS-v2-CLR and BNAS-v2-PC as follows. On one hand, the learning rate of architecture weights $\eta_{\alpha}$ is set to $3\times10^{-4}$ in BNAS-v2-CLR. Moreover, we employ CLR with $\beta=2$ to prevent the performance collapse issue of BNAS-v2. Beyond that, we set the batch size and learning rate to 256 and 0.1 for architecture search, respectively. On the other hand, $\eta_{\alpha}$ used for BNAS-v2-PC is set to $6\times10^{-4}$. The strategy of partial channel connections contributes BNAS-v2-PC to discover novel architecture with batch size 512 and learning rate 0.2, due to the contribution of memory efficiency.

For the architecture evaluation stage, BNAS-v2-CLR and BNAS-v2-PC use identical experimental settings as follows. BCNN is constructed by stacking 8 cells (2 deep cells and 1 broad cell in each convolution block, and 2 enhancement cells). We tune the number of initial input channels to fit the parameters following a mobile setting between 3$\sim$4M. Furthermore, BCNN is trained for 2000 epochs using SGD optimizer with batch size 128, initial learning rate 0.025 (the decayed way following the search phase), momentum 0.9, weight decay $3\times10^{-4}$. Moreover, cutout with 16 length \cite{devries2017improved} and drop path with a probability of 0.3 are adopted as previous NAS works \cite{liu2018darts,xu2019pc}. The experimental results ($mean \pm std$) are obtained by three repeated experiments.

\begin{table*}[!t]
\centering
\begin{threeparttable}[tbq]
\caption{Comparison of the proposed BNAS-v2 and other state-of-the-art NAS approaches on ImageNet}
\label{imagenet}
\begin{tabular}{lcccccc}
\hline
\multirow{2}{*}{\textbf{Architecture}} & \multicolumn{2}{c}{\textbf{Test Err. (\%)}} & \textbf{Params} & \textbf{Search Cost} & \textbf{Mult-Adds} & \multirow{2}{*}{\textbf{Topology}}\\
\cline{2-3}
& \textbf{top-1} & \textbf{top-5} & \textbf{(M)}  & \textbf{(GPU days)} & \textbf{(M)} \\

\hline
\hline
Inception-v1 \cite{szegedy2015going}  & 30.2 & 10.1 & 6.6   & -  & 1448 & deep\\
MobileNet \cite{howard2017mobilenets} & 29.4 & 10.5 & 4.2   & -  & 569  & deep\\
ShuffleNet \cite{zhang2018shufflenet} & 26.4 & 10.2 & $\sim$5    & -  & 524  & deep\\
\hline
\hline
AmoebaNet-A \cite{real2018regularized}& 25.5 & 8.0 & 5.1 & 3150 & 555 & deep\\
AmoebaNet-B \cite{real2018regularized}& 26.0 & 8.5 & 5.3 & 3150 & 555 & deep\\
AmoebaNet-C \cite{real2018regularized}& 24.3 & 7.6 & 6.4 & 3150 & 570 & deep\\
NASNet-A \cite{zoph2018learning}      & 26.0 & 8.4 & 5.3 & 1800 & 564 & deep\\
NASNet-B \cite{zoph2018learning}      & 27.2 & 8.7 & 5.3 & 1800 & 488 & deep\\
NASNet-C \cite{zoph2018learning}      & 27.5 & 9.0 & 4.9 & 1800 & 558 & deep\\
PNAS \cite{liu2018progressive}        & 25.8 & 8.1 & 5.1 & 225  & 588 & deep\\
DARTS (2nd order) \cite{liu2018darts}         & 26.7 & 8.7 & 4.7 & 1.50  & 574 & deep\\
ProxylessNAS (GPU) \citep{cai2018proxylessnas} $\dag$ & 24.9 & 7.5 & 7.1 & 8.30  & 465 & deep\\
SNAS (mild) \cite{xie2018snas}                & 27.3 & 9.2 & 4.3 & 1.50  & 522 & deep \\
P-DARTS (CIFAR-10) \cite{chen2019progressive} & 24.4 & 7.4  & 4.9 & 0.30 & 557 & deep\\
P-DARTS (CIFAR-100)\cite{chen2019progressive} & 24.7 &  7.5  & 5.1 & 0.30 & 577 & deep\\
PC-DARTS (CIFAR-10)\cite{xu2019pc}            & 25.1 &  7.8  & 5.3 & 0.10 & 586 & deep\\
PC-DARTS (ImageNet) \cite{xu2019pc}$\dag$     & \textbf{24.2} &  \textbf{7.3}  & 5.3 & 3.80 & 597 & deep\\
BNAS \cite{ding2020bnas}                      & 25.7 &  7.8  & 3.9 & 0.20 & - & broad\\
\hline
BNAS-v2-CLR-C2 (1st order) (ours)            & 27.3 & 9.0 & 4.4 & \textbf{0.09} & \textbf{441} & broad\\
BNAS-v2-CLR-C5 (1st order) (ours)            & 27.2 & 9.0 & \textbf{3.7} & \textbf{0.09} & 938 & broad\\
BNAS-v2-PC-C2 (2nd order) (CIFAR-10) (ours)      & 27.2 & 8.8 & 4.6 & \textbf{0.09} & 475 & broad\\
BNAS-v2-PC-C2 (1st order) (ImageNet) (ours)$\dag$            & 27.0 & 10.5   & 4.6 & \textbf{0.19}$\ddag$ & 576 & broad\\
\hline
\end{tabular}
\footnotesize
\begin{tablenotes}
\item[$\dag$] Those architectures are discovered on ImageNet directly.
\item[$\ddag$] State-of-the-art efficiency for proxyless architecture search on ImageNet.
\end{tablenotes}
\end{threeparttable}
\end{table*}

\subsubsection{Results and Analysis}

We visualize the best performing architectures learned by BNAS-v2-CLR and BNAS-v2-PC in Fig. \ref{cells}. Furthermore, TABLE \ref{tab::cifar10} summarizes the comparison of the proposed BNAS-v2 with other state-of-the-art NAS approaches.

The proposed BNAS-v2 takes full advantages of BCNN, so that great efficiency improvement can be obtained as shown in TABLE \ref{tab::cifar10}. Moreover, BNAS-v2 also achieves satisfactory accuracy. Particularly, BNAS-v2-PC (1st order) achieves 3.8x faster efficiency (state-of-the-art efficiency of 0.05 GPU days) and higher accuracy (i.e. 2.79 $\pm$ 0.19\% test error) than BNAS-CCE \cite{ding2020bnas} whose broad scalable architecture is identical with BNAS-v2. Compared with continuous relaxation-based DARTS \cite{liu2018darts}, the efficiency of BNAS-v2-CLR is 5x (0.09 GPU days) and 7.9x (0.19 GPU days) faster when using 1st and 2nd approximation orders, respectively. Beyond that, BNAS-v2-CLR also delivers higher accuracy due to the contribution of the proposed CLR. Compared with partial-connected PC-DARTS \cite{xu2019pc}, BNAS-v2-PC achieves state-of-the-art efficiency, 2x less computational cost (about 70 minutes) by using 1st order approximation. Beyond that, BNAS-v2-PC obtains competitive 2.79 $\pm$ 0.19 test error with 3.7M parameters. In particular, PC-DARTS suffers from the Out of Memory (OOM) issue when using 2nd order approximation with batch size 256 on a single NVIDIA GTX 1080Ti GPU (refer to TABLE \ref{tab::cifar10}). The reason is that there are no memory to construct an extra model for architecture search with 2nd order approximation in PC-DARTS (using about 12G memory with batch size 256). However, BNAS-v2-PC dose not suffer from the above OOM issue, due to enough memory is available for new model construction even though using batch size 512 for architecture search. Beyond that, due to the contribution of BCNN, all learned architectures achieve competitive even better performance just using 8 cells, instead of more than 17 cells. This characteristic implies that the architecture learned by BNAS-v2 excels in obtaining faster training and inference speed than those NAS approaches using deep scalable architecture for real-world applications, e.g. mobile devices.

\subsection{Results on ImageNet}

In this section, we do not only verify the transferability of the best performing architectures learned by BNAS-v2-CLR (1st order) and BNAS-v2-PC (2nd order), but also directly implement architecture search on ImageNet using BNAS-v2-PC (1st order).

\subsubsection{Experimental Settings for Transferability Verification}

In previous works using deep scalable architecture, the learned architecture is transferred to ImageNet by the following way. Three $3\times3$ convolutions with stride 2 are treat as the stem layers to reduce the resolution of input images from $224\times224$ to $28\times28$. Subsequently, the architecture learned on CIFAR-10 can be employed for image classification on ImageNet. Similarly, we leverage this way for the proposed approach named as BNAS-v2-C2 (e.g. BNAS-v2-CLR-C2, BNAS-v2-PC-C2 (CIFAR-10)), i.e. using 2 convolution blocks to construct the ImageNet classifier. Beyond that, another way for BCNN construction can be adopted. As described in BNAS \cite{ding2020bnas}, the number of convolution block is determined by the input size of the first convolution block, which is similar to the number of reduction cell in deep scalable architecture. Consequently, we also employ 5 convolution blocks to construct the ImageNet classifier named BNAS-v2-C5 (e.g. BNAS-v2-CLR-C5), for achieving possible better performance with full-scale representations. For BNAS-v2-C2, we set both the number of deep cell in each convolution block and enhancement cell to 2. For BNAS-v2-C5, there are 11 cells (1 deep cell and 1 broad cell in each convolution block, and 1 enhancement cell in each enhancement block) in the network. In the architecture evaluation phase of BNAS-v2-C5 on ImageNet, we train the architecture for 150 epochs with batch size 768 using 8 NVIDIA Tesla V100 GPUs. We also choose SGD as the optimizer with initial learning rate 0.1 (decayed by a factor of 0.1 at 80-th, 120-th and 140-th epoch), momentum 0.9 and weight decay $3\times10^{-5}$. Moreover, the initial input channels of BNAS-v2-C5 are set to 6, the label smoothing is set to 0.1, and the gradient clip bound is set to 5.0. Due to the topology difference, BNAS-v2-C2 can set larger batch size and learning rate than BNAS-v2-C5 for  training on ImageNet. Consequently, we leverage various hyper-parameters from BNAS-v2-C5 in terms of batch size, the initial value and decayed way of learning rate for BNAS-v2-C2 as follows. We train BNAS-v2-C2 for 250 epochs with batch size 1024 using 2 NVIDIA Tesla V100 GPUs. Similarly, we also choose the SGD optimizer with initial learning rate 0.5 whose decayed way is identical with PC-DARTS \cite{xu2019pc} used for ImageNet. Moreover, the initial input channels of BNAS-v2-CLR-C2 and BNAS-v2-PC-C2 (CIFAR-10) are set to 48 and 50, respectively.

\begin{figure}[t]
\centering
\includegraphics[width=0.48\textwidth]{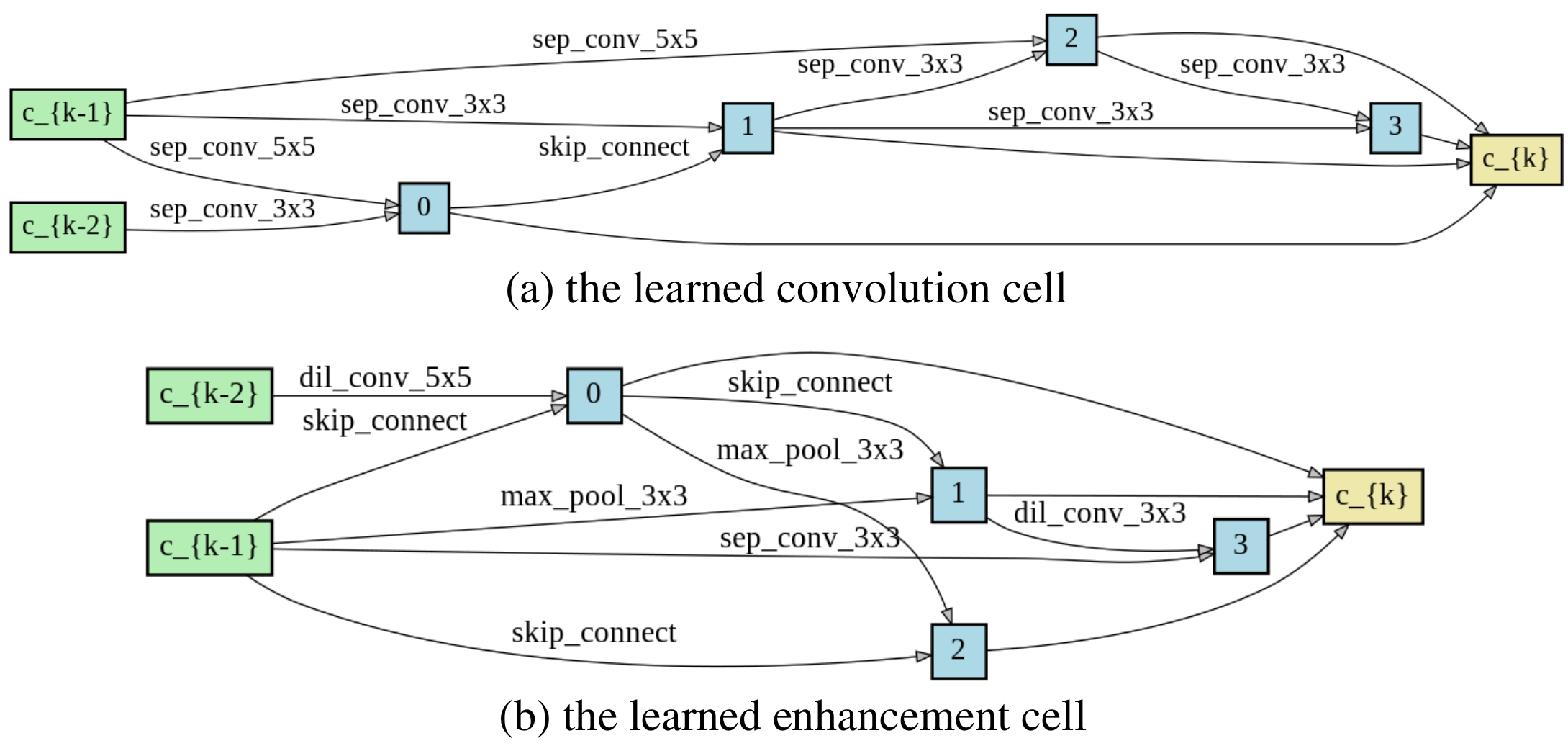}
\caption{The architecture learned by BNAS-v2-PC on ImageNet.}
\label{cells_imagenet}
\end{figure}

\subsubsection{Experimental Settings for Proxyless Architecture Search}

Two subsets contained 10\% and 2.5\% of 1.3 millions images, are randomly sampled from each category of training data of ImageNet for over-parameterized BCNN and architecture weights training, respectively. Similar to the search phase of PC-DARTS \cite{xu2019pc} on ImageNet, we treat three $3\times3$ convolutions with stride 2 as the stem layers to reduce the resolution of input images from $224\times224$ to $28\times28$. Subsequently, the over-parameterized BCNN used on CIFAR-10 can be employed for proxyless architecture search on ImageNet. Here, we use a single Tesla V100 GPU to discover architecture, and set the batch size and learning rate to 512 and 0.2, respectively. Other hyper-parameters are similar to BNAS-v2-PC used for architecture search on CIFAR-10. For the architecture directly learned on imageNet, we construct the image classifier with 2 convolution blocks by stacking 10 cells (3 deep cells in each convolution block, and a single enhancement block).

\subsubsection{Results and Analysis}

We visualize the convolution and enhancement cells learned by BNAS-v2-PC on ImageNet in Fig. \ref{cells_imagenet}, and summarize the comparison of BNAS-v2-PC (ImageNet) and other NAS approaches on ImageNet in TABLE \ref{imagenet}.

On one hand, both two BNAS-v2-CLR-based classifiers achieve competitive performance in terms of test error. Moreover, BNAS-v2-CLR-C5 achieves better accuracy with less parameters than BNAS-v2-CLR-C2. The above result implies that the strategy of multi-scale feature fusion contributes to the performance improvement of broad scalable architecture. Furthermore, the index of Mult-Adds of BNAS-v2-CLR-C2 is state-of-the-art. However, BNAS-v2-CLR-C5 has the largest index of Mult-Adds in TABLE \ref{imagenet} which dose not satisfy the mobile setting (i.e. smaller than 600M), even its index of parameters is just 3.7M. Here, the first three convolution blocks where the size of input is too large, lead to the above catastrophic phenomenon.

On the other hand, BNAS-v2-PC also delivers state-of-the-art search efficiency for proxyless architecture search on ImageNet. The entire search process spends around 4.6 hours (about 20x faster than state-of-the-art PC-DARTS) on a single NVIDIA Tesla V100 GPU. In addition, the indices of parameters and Mult-Adds of both two architectures learned by BNAS-v2-PC are better than vanilla PC-DARTS, especially for the later one. Beyond that, all architectures learned by BNAS-v2 achieve competitive accuracy just using few parameters, that verify the contribution of the multi-scale feature fusion of BCNN again. Moreover, BNAS-v2-PC-C2 (ImageNet) outperforms BNAS-v2-PC-C2 (CIFAR-10) in terms of the top-1 accuracy for large scale image classification. This indicates that the proposed BNAS-v2-PC is effective for proxyless architecture search on ImageNet.

Admittedly, there exists a performance gap for ImageNet classification between BNAS-v2 and PC-DARTS, i.e. broad and deep scalable architectures. Due to the restriction of computational resources, we can not determine a list of appropriate hyper-parameters for the learned broad architectures, e.g. the number of deep and enhancement cells. Beyond that, we argue that the number of convolution blocks plays an important role for broad deep scalable architecture in ImageNet classification task. This is similar to the backbone design of YOLOv3 \cite{redmon2018yolov3}, which obtains better detection performance using multi-scale prediction, i.e. feature maps with the sizes of $32\times32$, $16\times16$, and $8\times8$, than previous versions of YOLO. As a result, the optimal list of hyper-parameters including the number of convolution blocks, needs to be determined further through intensive experiments for large scale image classification. We believe that the above performance gap can be bridged with an optimal list of appropriate hyper-parameters in the future.

\subsection{Effectiveness of Confident Learning Rate}

\begin{figure*}[t]
\centering
\includegraphics[width=0.94\textwidth]{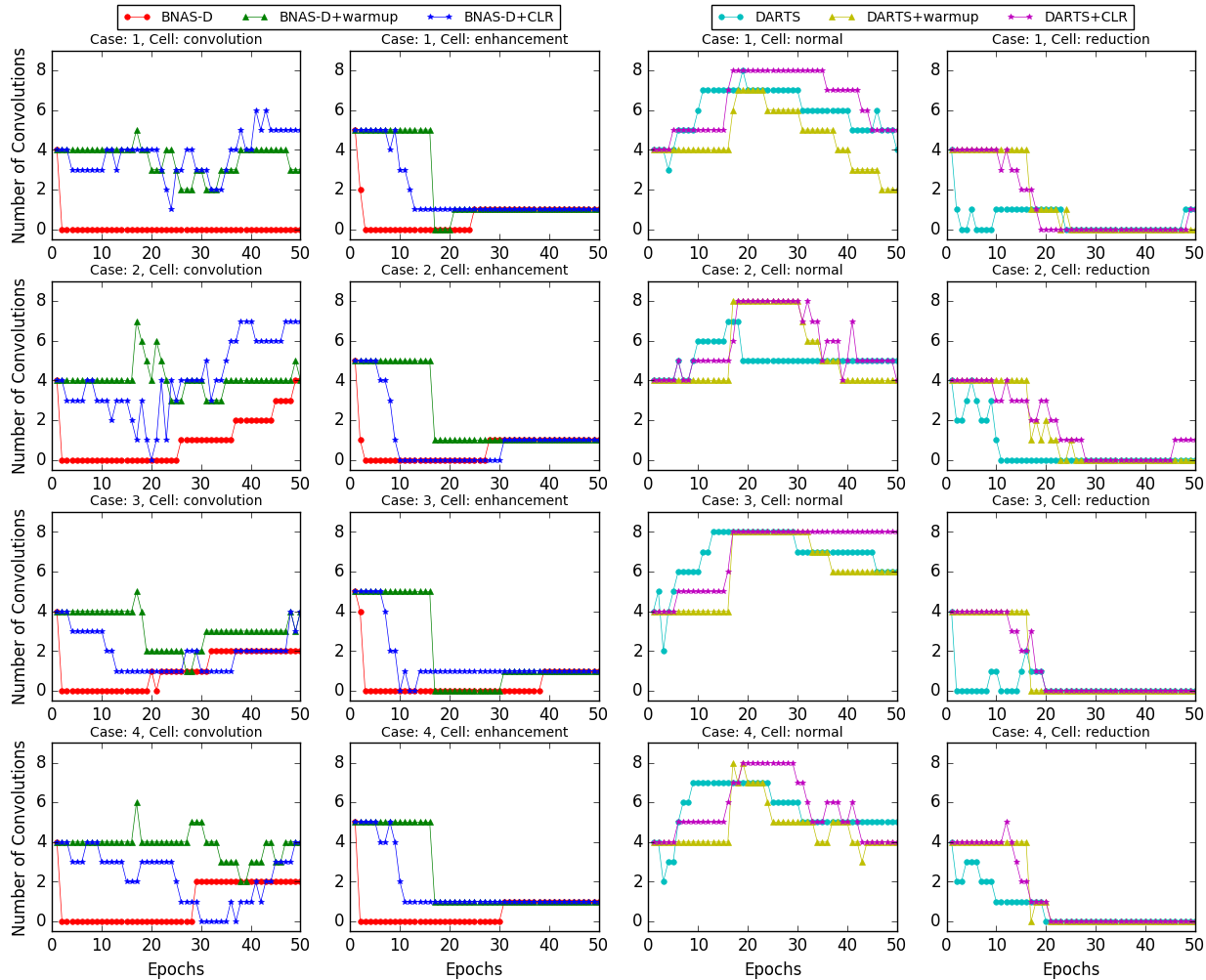}
\caption{The number of convolutions of entire search process in two cells learned by BNAS-D/DARTS, BNAS-D/DARTS with warmup and BNAS-D/DARTS with CLR under four repeated implementations where the confidence factor are set to 2 and 4 for BNAS-D and DARTS, respectively. Moreover, BNAS-D+CLR is identical with BNAS-v2-CLR. We use the sum of the number of convolution operations with regard to two cells as the index for performance evaluation. Best viewed in color.}
\label{effectiveness}
\end{figure*}

In this part, we implement two groups of ablation experiments with respect to BNAS-D and DARTS, to examine the effectiveness of the proposed CLR for performance collapse alleviation. To some extent, the number of convolutions in two types of cells, $\theta$, is basically proportional to the accuracy of model \cite{guo2020meets}, so that we employ $\theta$ as the index to examine the effectiveness of CLR for both BNAS-D and vanilla DARTS. Furthermore, we choose warmup strategy as the comparative method to further verify the effectiveness of CLR. We repeat running six methods, vanilla BNAS-D/DARTS, BNAS-D/DARTS with warmup, and BNAS-D/DARTS with CLR, for four times, and show the number of convolutions of entire search process in Fig. \ref{effectiveness}.

For vanilla performance-collapse-suffered BNAS-D, the mean value of $\theta$ is about 3 which tends to deliver poor performance. Moreover, we train the architecture of case 2 (most convolution operations in the learned architecture over four cases) learned by vanilla BNAS-D for 600 epochs. The architecture achieves 3.26\% test error with 3.6M parameters under default hyper-parameters of BNAS-v2 on CIFAR-10. The warmup strategy contributes to the performance collapse issue mitigation of BNAS-D in term of $\theta$, where the mean value of $\theta$ is about 4.5. For the proposed CLR, $\theta$ of four repeated implementations are 6, 8, 5 and 5, respectively. Obviously, there are more convolutions in the learned cells by BNAS-D with CLR than warmup. Moreover, we also train the architecture of case 1 learned by BNAS-D using CLR for 600 epochs. The architecture achieves 2.90\% test error with 3.4M parameters under the above settings for architecture evaluation.

%\begin{table}[!t]
%\centering
%\caption{The examination of transferable ability between broad and deep scalable architectures}
%\label{Poor Transferable Ability}
%\begin{tabular}{lccc}
%\hline
%\multirow{2}{*}{\textbf{Architecture}} & \textbf{Number}   & \textbf{Test Error} & \textbf{Parameters} \\
% & \textbf{of Cells} & \textbf{(\%)} & \textbf{(M)} \\
%\hline
% \multirow{5}{*}{B-DARTS}
% & 8   & 3.39 & 3.3 \\
% & 11  & 3.45 & 3.4 \\
% & 14  & 3.27 & 3.4 \\
% & 17  & 3.60 & 3.5 \\
% & 20  & 3.83 & 3.4 \\
%\hline
%\hline
% \multirow{5}{*}{B-PC-DARTS}
% & 8   & 3.14 & 3.7 \\
% & 11  & 3.04 & 3.6 \\
% & 14  & 3.17 & 3.6 \\
% & 17  & 3.51 & 3.7 \\
% & 20  & 3.91 & 3.5 \\
%\hline
%\end{tabular}
%\end{table}

From the overall perspective, $\theta$ for BNAS-D with CLR (i.e. BNAS-v2-CLR) decreases firstly, and then increases to a high value. In the decreasing phase of $\theta$, those weight-free operations are prone to be equipped with larger weights than the weight-equipped one, although the confident learning rate is small enough. The reason of above situation is that the outputs of weight-free operations are more consistent with its input, which is preferred by gradient-based search algorithm \cite{xu2019pc}. With the convergence of over-parameterized BCNN, the increasing phase of $\theta$ begins. Those well-optimized weight-equipped operations contribute to the improvement of validation accuracy, so that the search strategy starts to prefer them.

Different from BNAS-D, we set the value of confidence factor $\beta$ to 4 rather than 2. Through extensive experiments, we draw a conclusion that \emph{deep scalable architecture should be equipped with larger value of $\beta$ than the broad one, for training the over-parameterized model well}. For vanilla DARTS, the mean value of $\theta$ is about 5, and only the first implementation learns to utilize a convolutional operation in reduction cell. Here, a convolution in reduction cell tends to achieve satisfactory performance with larger probability than those cells consisted of all weight-free operations. Beyond that, DARTS with warmup delivers equal or worse performance than vanilla DARTS in term of the number of convolutional operations. Moreover, it can not learn to employ convolutional operation in reduction cell for possible improvement. DARTS with CLR learns 6, 5, 8, 4 convolutions under four repeated implementations, whose mean value is about 6 (1 larger than vanilla DARTS). Importantly, the proposed CLR excels in discovering the reduction cell of DARTS with convolutional operation for potential performance improvement.

\section{Conclusions}
\label{conclusions}

BNAS can not take full advantages of BCNN for architecture search, due to the influence of unfair training issue. In order to improve the efficiency of BNAS further, we propose BNAS-v2. Particularly, we employ the strategy of continuous relaxation to mitigate the unfair training issue in BNAS. However, a consequent issue of continuous relaxation named performance collapse arises. For this, we provide two solutions in BNAS-v2, i.e. BNAS-v2-CLR and BNAS-v2-PC. On one hand, we propose Confident Learning Rate (CLR) in BNAS-v2-CLR, that considers the confidence of gradient for architecture weights update increasing with the training time of over-parameterized BCNN, to prevent the issue of performance collapse. On the other hand, for BNAS-v2-PC, we introduce the combination of partial channel connections and edge normalization that can not only prevent the performance issue, but also make BNAS-v2-PC more memory-efficient than BNAS-v2-CLR.

The proposed BNAS-v2 delivers state-of-the-art efficiency on both CIFAR-10 and ImageNet. Particularly, BNAS-v2-PC achieves 3.8x faster efficiency (state-of-the-art efficiency of 0.05 GPU days) and higher accuracy (i.e. 2.79 $\pm$ 0.19\% test error) on CIFAR-10 than BNAS-CCE whose broad scalable architecture is identical with BNAS-v2. Furthermore, BNAS-v2 also can realize the proxyless architecture search on ImageNet with state-of-the-art efficiency of 0.19 GPU days. Beyond that, the proposed CLR contributes both BNAS-v2 and DARTS to learn more convolutions in two types of cells for performance improvement. However, BNAS-v2 dose not always improve the efficiency without performance degradation, especially for large scale image classification task. We will utilize knowledge distillation \cite{heo2019knowledge} to solve this issue in the future.

\ifCLASSOPTIONcaptionsoff
  \newpage
\fi

\bibliographystyle{IEEEtranN}
\bibliography{mybibfile}

%\end{IEEEbiography}

\end{document}